\documentclass[letterpaper, 10 pt, conference]{ieeeconf}  % Comment this line out if you need a4paper

\IEEEoverridecommandlockouts                              % This command is only needed if 
\usepackage{amsmath, amsfonts, amssymb}
\usepackage{xcolor}
\usepackage{graphicx}
\usepackage{multirow}

\usepackage{algorithm,algcompatible}
\algnewcommand\INPUT{\item[\textbf{Input:}]}%
\algnewcommand\OUTPUT{\item[\textbf{Output:}]}%

\newcommand{\bni}[1]{\boldsymbol{\mathrm{#1}}} % bold no italic symbol

\title{\LARGE \bf
A Tightly Coupled LiDAR-IMU Odometry through\\ Iterated Point-Level Undistortion
}

\author{Keke Liu$^{1}$, Hao Ma$^{2}$ and Zemin Wang$^{3}$% <-this % stops a space
\thanks{$^{1}$Keke Liu
{\tt\small kekeliu@whu.edu.cn}}%
\thanks{$^{2}$Hao Ma
{\tt\small mahao\_fido@whu.edu.cn}}%
\thanks{$^{3}$Zemin Wang
{\tt\small zeminwang@whu.edu.cn}}%
}

\begin{document}

\maketitle
\thispagestyle{empty}
\pagestyle{empty}

\begin{abstract}

Scan undistortion is a key module for LiDAR odometry in high dynamic environment with high rotation and translation speed. The existing line of studies mostly focuses on one pass undistortion, which means undistortion for each point is conducted only once in the whole LiDAR-IMU odometry pipeline. In this paper, we propose an optimization based tightly coupled LiDAR-IMU odometry addressing iterated point-level undistortion. By jointly minimizing the cost derived from LiDAR and IMU measurements, our LiDAR-IMU odometry method performs more accurate and robust in high dynamic environment. Besides, the method characters good computation efficiency by limiting the quantity of parameters.

\end{abstract}

\section{Introduction}
\label{sec:intro}

Most consumer-grade LiDAR has mechanical rotating structure inside. A LiDAR scan within a sweep period unavoidably suffers from distortion, because a surrounding LiDAR scan is accumulatively acquired in one rotation period instead of being taken simultaneously. On stable carriers with smooth movement, such as trolley or car, LiDAR only odometry \cite{2014LOAM} is enough to perform scan undistortion based on the rigid linear motion assumption. However, when the LiDAR carrier move rapidly with high rotation and translation speed, scan distortion could lead to decimeter-level localization and mapping error. IMU is usually fused with LiDAR to achieve more accurate and robust poses.

There are several methods relating to the LiDAR-IMU fusion odometry. Most of the methods are loosely coupled or scan-level tightly coupled, with a typical straightforward odometry pipeline, which can be simplified as "undistort scan by IMU $\rightarrow$ extract feature points from scan $\rightarrow$ match scan with map $\rightarrow$ fuse LiDAR and IMU to get scan pose". Apart from slightly different details between these methods, they all have only one pass scan undistortion by IMU measurements. In \cite{liosam2020shan}, scan undistortion relies on IMU angular velocity integration only and translation is ignored by setting velocity to zero. \cite{ye2019tightly} uses IMU preintegration to do scan undistortion, considering both rotation and translation.

As far as we know, point-level tightly coupled methods are scarce. \cite{xu2021fast} takes each point's error and covariance into consideration by an iterative kalman filter method, however it is still a one pass undistortion method. In this paper, we propose an optimization based tightly coupled LiDAR-IMU odometry addressing iterated point-level undistortion.

\section{Method}
\label{sec:method}

\subsection{LiDAR-IMU odometry overview}
\begin{figure}[h!]
	\begin{center}
		\includegraphics[width=\linewidth]{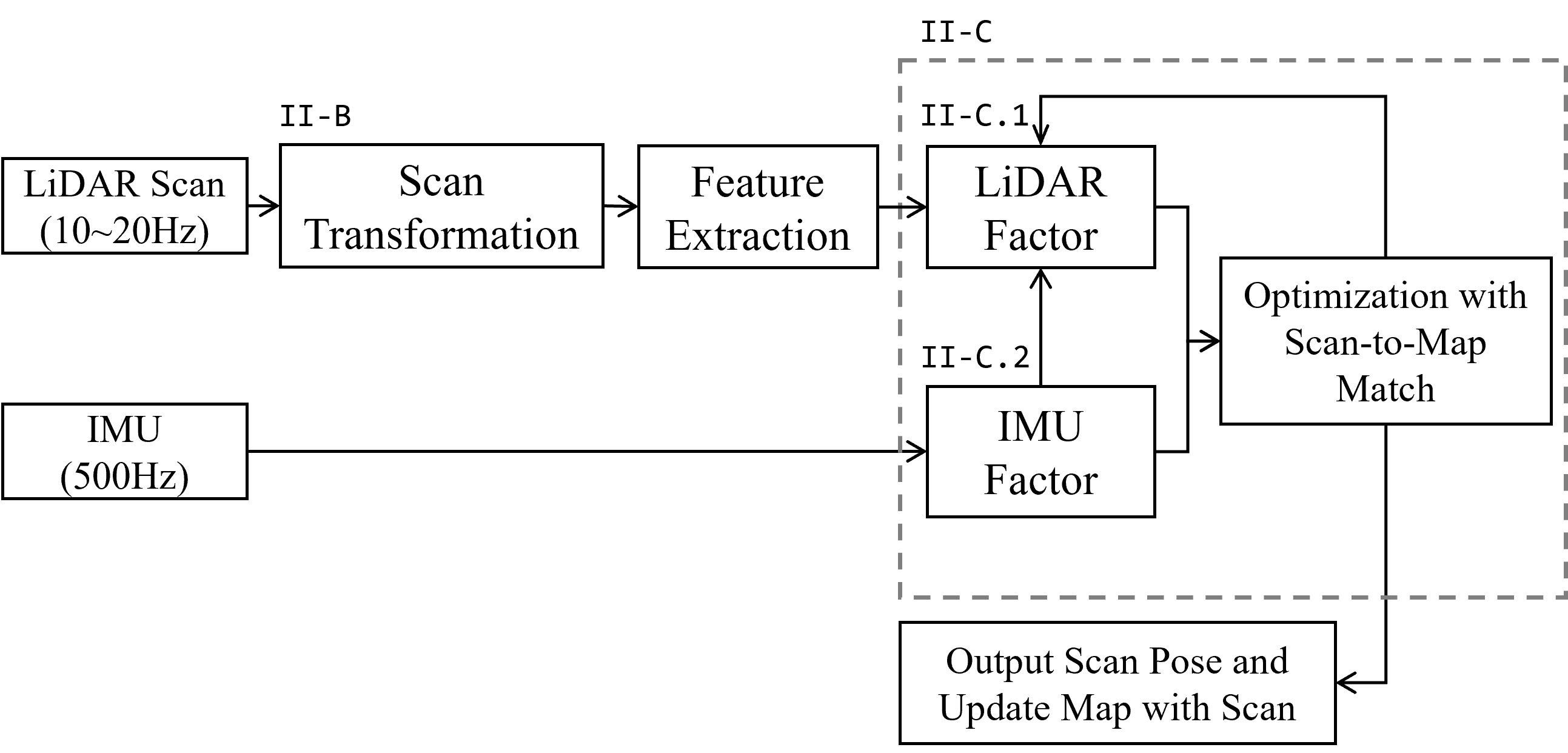}
	\end{center}
	\caption{LiDAR-IMU odometry pipeline overview.}
	\label{fig:system_overview}
\end{figure}

Fig. \ref{fig:system_overview} illustrates a pipeline overview  of our proposed LiDAR-IMU odometry. IMU data with high frequency up to 500Hz is processed to \textbf{IMU factor} by IMU preintegration in \ref{sec:imu_factor}. The LiDAR collects a scan with surrounding point cloud in a much slower frequency. ``Scan Transformation'' transforms each scan from LiDAR frame to IMU frame, aiming to ignore the notations of extrinsic parameters (specific explanation in \ref{sec:notations}). ``Feature Extraction'' extracts feature points on sharp edges and planar surfaces by methods described in \cite{2014LOAM}. Then feature points are constructed to \textbf{LiDAR factor} with point-level undistortion in \ref{sec:lidar_factor}. IMU factor and LiDAR factor will be collected into optimization problem to optimize out kinetic state of the new scan.

\subsection{Notations}
\label{sec:notations}
\begin{table}[H]
	\caption{\label{tab:notations}Notations in this paper.}
	\centering
	\setlength{\tabcolsep}{1mm}
	\begin{tabular}{c|l}
		\hline
		Notation                 & Description                                              \\
		\hline
		$t_k$                    & End time of the scan $k$ or start time of the scan $k+1$ \\
		$t_i$                    & Time of the IMU data $i$                                 \\
		$t_j$                    & Time of the feature point $j$ in a scan                  \\
		$p_{b_k}^w$, $q_{b_k}^w$ & Transformation from LiDAR frame $k$ to world frame       \\
		$p_j^{b_j}$              & Raw measurements of point $j$ in a scan                  \\
		\hline
	\end{tabular}
\end{table}
Table.~\ref{tab:notations} describes some significant notations in this paper. It is noteworthy that LiDAR is rigidly binded to IMU and the extrinsic parameters between them has been calibrated well before odometry pipeline. Accordingly, we introduce the step ``Scan Transformation'' to transform LiDAR scan points from LiDAR frame into IMU frame. After ``Scan Transformation'', the LiDAR frame and IMU frame can be treated identical and the extrinsic parameter notations between them will be ignored in this paper.

\subsection{State Estimation}
\subsubsection{IMU Factor}
\label{sec:imu_factor}

IMU can get raw angular velocity and linear acceleration of rigid body with measurement model as follows:
\begin{equation}
	\begin{split}
		\hat{a}_t &= a_t + R_w^t g^w + b_{a_t} + n_a\\
		&= R_w^t (a_t^w+g^w)+b_{a_t}+n_{a} \\
		\hat{\omega}_t &= \omega_t+b_{\omega_t}+n_\omega
	\end{split}
\end{equation}
where $\hat{\omega}$ and $\hat{a}$ is the raw gyroscope and accelerometer measurements, respectively. The IMU measurements is effected by bias, white noise and random walk noise. The white noise and random walk noise satisfy the Gaussian distribution:
\begin{equation}
	\begin{gathered}
		n_a \sim \mathcal{N}(\boldsymbol{0},\boldsymbol{\sigma}_a^2) \\
		n_\omega \sim \mathcal{N}(\boldsymbol{0},\boldsymbol{\sigma}_\omega^2) \\
		\dot{b}_{a_t} = n_{b_a} \sim \mathcal{N}(\boldsymbol{0},\boldsymbol{\sigma}_{b_a}^2) \\
		\dot{b}_{\omega_t} = n_{b_{\omega}} \sim \mathcal{N}(\boldsymbol{0},\boldsymbol{\sigma}_{b_{\omega}}^2) \\
	\end{gathered}
\end{equation}

In world frame, kinetic equations can be written as follows:
\begin{equation}
	\begin{split}
		p_{b_k}^w & = p_{b_{k-1}}^w + v_{b_{k-1}}^w \Delta t_k \\
		&\quad\quad + \iint_{t_{k-1}}^{t_k} (R^w_t(\hat{a}_t-b_{a_t}-n_a)-g^w) dt^2 \\
		v_{b_k}^w & = v_{b_{k-1}}^w +
		\int_{t_{k-1}}^{t_k}    (R^w_t(\hat{a}_t-b_{a_t}-n_a)-g^w) dt \\
		q_{b_k}^w & = q_{b_{k-1}}^w \otimes
		\int_{t_{k-1}}^{t_k}    \frac{1}{2} \Omega(\hat{w}_t-b_{\omega_t}-n_\omega)q_t^{b_{k-1}} dt
	\end{split}
\end{equation}
where $\hat{a}_t$ is in IMU frame and $a_t^w$ is in world frame and
\begin{equation}
	\Omega(\omega)=\left[\begin{array}{cc}
			-\omega^\wedge & \omega \\
			-\omega^T      & 0
		\end{array}\right],
	\omega^\wedge=\left[\begin{array}{ccc}
			0         & -\omega_z & \omega_y  \\
			\omega_z  & 0         & -\omega_x \\
			-\omega_y & \omega_x  & 0
		\end{array}\right]
\end{equation}

Instead of traditional forward propagation for IMU preintegration in \cite{forster2016manifold} and \cite{qin2018vins}, we introduce backward propagation to unify the derivation formulas in \ref{sec:imu_factor} and \ref{sec:lidar_factor}:
\begin{equation}
	\begin{split}
		p_{b_{k-1}}^w & = p_{b_k}^w - v_{b_{k-1}}^w \Delta t_k \\
		&\quad\quad + \iint_{t_k}^{t_{k-1}} (R^w_t(\hat{a}_t-b_{a_t}-n_a)-g^w) dt^2 \\
		v_{b_{k-1}}^w & = v_{b_k}^w +
		\int_{t_k}^{t_{k-1}}     (R^w_t(\hat{a}_t-b_{a_t}-n_a)-g^w) dt \\
		q_{b_{k-1}}^w & = q_{b_k}^w \otimes
		\int_{t_k}^{t_{k-1}}     \frac{1}{2} \Omega(\hat{w}_t-b_{\omega_t}-n_\omega)q_t^{b_k} dt
	\end{split}
\end{equation}

In IMU frame w.r.t. previous time,
\begin{equation} \label{eq:vins-preintegration}
	\begin{split}
		R_w^{b_k}p_{b_{k-1}}^w &= R_w^{b_k}( p_{b_k}^w - v_{b_k}^w\Delta{t} + \frac12g^w\Delta{t}^2) + \alpha_{b_{k-1}}^{b_k} \\
		R_w^{b_k}v_{b_{k-1}}^w &= R_w^{b_k}( v_{b_k}^w + g^w\Delta{t} ) + \beta_{b_{k-1}}^{b_k} \\
		q_w^{b_k} \otimes q_{b_{k-1}}^w &= \gamma_{b_{k-1}}^{b_k}
	\end{split}
\end{equation}

where contains preintegration quantity $\alpha_{b_{k-1}}^{b_k}$, $\beta_{b_{k-1}}^{b_k}$ and $\gamma_{b_{k-1}}^{b_k}$.
\begin{equation}
	\begin{split}
		\alpha_{b_{k-1}}^{b_k} &= \iint_{t_k}^{t_{k-1}} R_t^{b_k}(\hat{a}_t-b_{a_t}-n_a)dt^2 \\
		\beta_{b_{k-1}}^{b_k} &= \int_{t_k}^{t_{k-1}} R_t^{b_k}(\hat{a}_t-b_{a_t}-n_a)dt \\
		\gamma_{b_{k-1}}^{b_k} &= \int_{t_k}^{t_{k-1}} \frac12\Omega(\hat{\omega}_t-b_{\omega_t}-n_\omega) \gamma_t^{b_k} dt
	\end{split}
\end{equation}
From above equations, we can see preintegration quantity is only related with IMU bias $b_{a_t}$ and $b_{\omega_t}$.

In discrete time, preintegration quantity can be iterated as Eq. \ref{eq:discrete-time-preintegration} with mid-point integration. Noise $n_{a_t}$ and $n_{g_t}$ are ignored in the equation because they can not be predicted and their integration can be treated as zero during small interval. $\alpha_i^{b_k}$, $\beta_i^{b_k}$, $\gamma_i^{b_k}$ will be initialized to be $\boldsymbol{0}$, $\boldsymbol{0}$ and identity rotation respectively. Furthermore, IMU bias between consecutive frame changes quite small so it can be treated as constant between frame $i$ and frame $i-1$.
\begin{equation} \label{eq:discrete-time-preintegration}
	\begin{split}
		\hat{\alpha}_{i-1}^{b_k} &= \hat{\alpha}_i^{b_k} -
		\frac12(\hat{\beta}_i^{b_k}+\hat{\beta}_{i-1}^{b_k})\delta{t} \\
		\hat{\beta}_{i-1}^{b_k} &= \hat{\beta}_i^{b_k} - \frac12\left[\hat{\gamma}_i^{b_k}(\hat{a}_i-b_{a_i})+\hat{\gamma}_{i-1}^{b_k}(\hat{a}_{i-1}-b_{a_i})\right]\delta{t} \\
		\hat{\gamma}_{i-1}^{b_k} &= \hat{\gamma}_i^{b_k} \otimes
		\begin{bmatrix}
			1 \\ -\frac12(\frac{\hat{w}_i+\hat{w}_{i-1}}2-b_{\omega_i})\delta{t}
		\end{bmatrix}
	\end{split}
\end{equation}

Applying Eq.~\ref{eq:discrete-time-preintegration} we get IMU factor as following:
\begin{equation}\label{eq:imu_factor}
	r_\mathcal{B}(\bni{x}_k) =
	\begin{bmatrix}
		R_w^{b_k}( p_{b_k}^w - v_{b_k}^w\Delta{t} + \frac12g^w\Delta{t}^2 - p_{b_{k-1}}^w) + \alpha_{b_{k-1}}^{b_k} \\
		R_w^{b_k}( v_{b_k}^w + g^w\Delta{t} - v_{b_{k-1}}^w) + \beta_{b_{k-1}}^{b_k}                                \\
		2\left[ q_w^{b_k} \otimes q_{b_{k-1}}^w \otimes (\gamma_{b_{k-1}}^{b_k})^{-1} \right]_{xyz}                 \\
		b_{a_k} - b_{a_{k-1}}                                                                                       \\
		b_{\omega_k} - b_{\omega_{k-1}}
	\end{bmatrix}
\end{equation}
Derivation of Jacobian and covariance matrix is similar with \cite{qin2018vins}.

\subsubsection{LiDAR factor with point-level undistortion}
\label{sec:lidar_factor}

From Eq.~\ref{eq:vins-preintegration}, we can derive a priori point undistortion equation for point $j$ only through IMU preintegration:
\begin{equation}
	\label{eq:point_undistortion}
	\begin{split}
		\bar{p} _{b_j}^{b_k}
		&= R_w^{b_k}p_{b_j}^w + p_w^{b_k} \\
		&= R_w^{b_k}( p_{b_k}^w - v_{b_k}^w\Delta{t} + \frac12g^w\Delta{t}^2) + \alpha_{b_j}^{b_k} + p_w^{b_k} \\
		&= R_w^{b_k}( -v_{b_k}^w\Delta{t} + \frac12g^w\Delta{t}^2) + \alpha_{b_j}^{b_k} \\
		\bar{q} _{b_j}^{b_k} &= \gamma_{b_j}^{b_k}
	\end{split}
\end{equation}

The relative pose estimated by ``LiDAR Scan-To-Map Match'' between $t_{k-1}$ and $t_k$ in one pass is often viewed as a reliable one, thus we try to align the relative pose estimated by IMU to that of LIDAR. To achieve this, correction on point $j$ is introduced to linearly interpolate among the estimated relative poses from IMU. The goal of correction is to force the the two relative poses between $t_{k-1}$ and $t_k$ being the same. After the correction, relative poses of IMU are updated and LiDAR's are further optimized. According to above inference, we design the following correction equation for point $j$:

\begin{equation}
	\label{eq:perturbation_delta_pq}
	\begin{split}
		\delta T_j &= slerp \left(I, ({T_{b_k}^w})^{-1} T_{b_{k-1}}^w ({{}\bar{T} _{b_{k-1}}^{b_k}})^{-1}, \mu_j \right) \\
		&= \{ \delta p_j, \delta q_j \} \\
		&=
		\{
		\mu_j(R_w^{b_k}(R_{b_{k-1}}^w \bar{p}_{b_k}^{b_{k-1}} +p_{b_{k-1}}^w) +p_w^{b_k}), \\
		&\ \ \ \ \ \ \
		2\mu_j \left[ q_w^{b_k} \otimes q_{b_{k-1}}^w \otimes (\gamma_{b_{k-1}}^{b_k})^{-1} \right]_{xyz}
		\}
	\end{split}
\end{equation}
where $\mu_j = (t_k-t_j)/(t_{k}-t_{k-1})$ is the linear interpolation factor under the assumption that the slight correction can be linearly applied to each point, $\bar{T}_{b_{k-1}}^{b_k} = \{\bar{p}  _{b_{k-1}}^{b_k}, \bar{q} _{b_{k-1}}^{b_k} \}$ is the priori relative pose between frame $k-1$ and frame $k$ via IMU preintegration.

Through left multiplying Eq.~\ref{eq:perturbation_delta_pq} to Eq.~\ref{eq:point_undistortion}, we can get point undistortion equation for point $j$ with correction:
\begin{equation}
	\label{eq:point_undistortion_corrected}
	\begin{split}
		\check{p} _{b_j}^{b_k}
		&= \delta R_j \bar{p}_{b_j}^{b_k} + \delta{p_j} \\
		\check{q} _{b_j}^{b_k} &= \delta{q_j} \otimes \bar{q}_{q_j}^{b_k}
	\end{split}
\end{equation}

Applying Eq.~\ref{eq:point_undistortion_corrected} to the point-line distance minimization equation from paper~\cite{2014LOAM}, we can get:
\begin{equation}\label{eq:lidar_factor1}
	\begin{split}
		d_{\epsilon}
		&= n^\wedge (R_{b_k}^w (\check{R} _{b_j}^{b_k}p_j^{b_j} + \check{p} _{b_j}^{b_k}) + p_{b_k}^w - p_0) \\
		% \frac{\partial d_{\epsilon}}
		% {\partial[p_{b_k}^w,q_{b_k}^w,v_{b_k}^w]} &=
		% n^\wedge
		% \begin{bmatrix}
		% 	I & -R_{b_k}^w (\gamma_{b_j}^{b_k}p_i^{b_j} + \alpha_{b_j}^{b_k})^\wedge & I\Delta{t}
		% \end{bmatrix}
	\end{split}
\end{equation}
Here, scan point $p_{j}^{b_j}$ corresponds to the line which is represented by a direction vector $n$ and one on-line point $p_0$.

Applying Eq.~\ref{eq:point_undistortion_corrected} to the point-plane distance minimization equation from paper~\cite{2014LOAM}, we can get:
\begin{equation}\label{eq:lidar_factor2}
	\begin{split}
		d_{\Pi}
		&= n^T (R_{b_k}^w (\check{R} _{b_j}^{b_k}p_j^{b_j} + \check{p} _{b_j}^{b_k}) + p_{b_k}^w - p_0) \\
		% \frac{\partial d_{\epsilon}}
		% {\partial[p_{b_k}^w,q_{b_k}^w,v_{b_k}^w]} &=
		% n^\wedge
		% \begin{bmatrix}
		% 	I & -R_{b_k}^w (\gamma_{b_j}^{b_k}p_i^{b_j} + \alpha_{b_j}^{b_k})^\wedge & I\Delta{t}
		% \end{bmatrix}
	\end{split}
\end{equation}
Here, scan point $p_{j}^{b_j}$ corresponds to the plane which is represented by a norm vector $n$ and one on-plane point $p_0$.

The jacobian of Eq.~\ref{eq:lidar_factor1} and Eq.~\ref{eq:lidar_factor2} w.r.t. state $\bni{x}_k$ is complicated and can derived as follows:
\begin{equation}
	\begin{split}
		\frac{\partial f(x)g(x)v}{\partial\delta x}
		&= \lim_{\delta x\to0} \frac{f(x\oplus\delta x)g(x\oplus\delta x)v-f(x)g(x)v}{\delta x} \\
		&= \lim_{\delta x\to0} \frac{f(x\oplus\delta x)g(x)v - f(x)g(x)v}{\delta x} \\
		&\ \ \ \ + \lim_{\delta x\to0} \frac{f(x)g(x\oplus\delta x)v - f(x)g(x)v}{\delta x} \\
		&= \tilde{f}(x)g(x)v + f(x)\tilde{g}(x)v
	\end{split}
\end{equation}
where $\delta x$ means the right perturbation on manifold and the $\tilde{f}(x)$ means derivative of $f(x)$ w.r.t. $\delta x$.

\subsubsection{State optimization}
The full state vector of the new scan is defined as:
\begin{equation}
	\begin{split}
		\bni{x}_k = [p_{b_k}^w, v_{b_k}^w, q_{b_k}^w, b_{a_k}, b_{\omega_k}]
	\end{split}
\end{equation}
We minimize the sum of prior and the Mahalanobis norm of all measurement residuals to obtain a maximum posteriori estimation:
\begin{equation}\label{eq:optimization_problem}
	\min_{\bni{x}_k}
	\left\{
	\left\lVert
	r_\mathcal{B}(\bni{x}_k)
	\right\rVert _{P_{b_{k-1}}^{b_k}} ^2
	+
	\sum _{p_j^{b_j}\in\mathcal{C}}
	\left\lVert
	r_\mathcal{C}(\bni{x}_k, p_j^{b_j})
	\right\rVert _{P_j} ^2
	\right\}
\end{equation}
where $r_\mathcal{C}(\bni{x}_k, p_j^{b_j})$ is the residual described in \ref{sec:lidar_factor}.

The algorithm of pose estimation with point-level undistortion can be described in Algorithm.~\ref{algo:pose_estimation}.
\begin{algorithm}
	\caption{\label{algo:pose_estimation}Pose estimation with point-level undistortion}
	\begin{algorithmic}[1]
		\REQUIRE Global map $\mathcal{M}$, State $\bni{x}_{b_{k-1}}^w$ for scan $k-1$.
		\INPUT Feature cloud $\mathcal{C}$ of scan $k$, IMU measurements $\mathcal{I}$ from $t_{k-1}$ to $t_k$.
		\OUTPUT State $\bni{x}_{b_k}^w$ for scan $k$.
		\STATE Construct IMU factor by Eq.~\ref{eq:imu_factor}
		\FORALL{$p_j^{b_j}\in\mathcal{C}$}
		\STATE Find corresponding edge or planar in global map $\mathcal{M}$
		\STATE Construct LiDAR factor by Eq.~\ref{eq:lidar_factor1} or Eq.~\ref{eq:lidar_factor2}
		\ENDFOR
		\STATE Construct optimization problem by Eq.~\ref{eq:optimization_problem}
		\STATE Solve Eq.~\ref{eq:optimization_problem} to get $\bni{x}_{b_k}^w$
	\end{algorithmic}
\end{algorithm}

\subsection{Initialization of gravity vector and bias}
During initialization, device need to be set still for 10 seconds to initialize the gravity vector and gyroscope bias. Readers are recommended to refer to the paper \cite{qin2018vins} for more details.

% \section{Results and Evaluation}
% \label{sec:results}

%\section{Conclusion}
\section{Conclusion and Future Work}
In this paper, we novelly propose a LiDAR-IMU odometry with iterated point-level undistortion in a tightly coupled way. Different from previous one pass undistortion methodologies, we try to align the relative pose estimated by IMU to that via LiDAR to achieve better undistortion performance. Theoretically, our method is highly applicable in high dynamic environment with high rotation and translation speed. In the future, we need to collect data to do experiments to validate the effectiveness of our method.

% \newpage
\bibliographystyle{IEEEtran}
\bibliography{root_ref}

\end{document}